\documentclass{article}
\usepackage{spconf,graphicx}
\usepackage{hyperref}
\usepackage{floatrow}
\usepackage{amsmath} 
\usepackage{amssymb} 
\usepackage{float}

\title{Adv-KD: Adversarial Knowledge Distillation for Faster Diffusion Sampling}

\name{Kidist Amde Mekonnen$^{\dagger}$, Nicola Dall'Asen\textsuperscript{*} , Paolo Rota\textsuperscript{*}}
\address {University of Trento, Trento , Italy \textsuperscript{*}$^{\dagger}$ \\
    $^{\dagger}${kidistamdie@gmail.com}\\
        \textsuperscript{*}{\{nicola.dallasen, paolo.rota\}@unitn.it}}

\begin{document}
\ninept
\maketitle

\begin{abstract}

Diffusion Probabilistic Models (DPMs) have emerged as a powerful class of deep generative models, achieving remarkable performance in image synthesis tasks. However, these models face challenges in terms of widespread adoption due to their reliance on sequential denoising steps during sample generation. This dependence leads to substantial computational requirements, making them unsuitable for resource-constrained or real-time processing systems. To address these challenges, we propose a novel method that integrates denoising phases directly into the model's architecture, thereby reducing the need for resource-intensive computations. 

Our approach combines diffusion models with generative adversarial networks (GANs) through knowledge distillation, enabling more efficient training and evaluation. By utilizing a pre-trained diffusion model as a teacher model, we train a student model through adversarial learning, employing layerwise transformations for denoising and submodules for predicting the teacher model's output at various points in time. This integration significantly reduces the number of parameters and denoising steps required, leading to improved sampling speed at test time.

We validate our method with extensive experiments, demonstrating comparable performance with reduced computational requirements compared to existing approaches. By enabling the deployment of diffusion models on resource-constrained devices, our research mitigates their computational burden and paves the way for wider accessibility and practical use across the research community and end-users.

Our code is publicly available at my  \href{https://github.com/kidist-amde/Adv-KD}{github-page}
\end{abstract}
\section{Introduction}
\label{sec:intro}
In recent years, a class of deep generative models known as diffusion models \cite{sohl2015deep,ho2020denoising,song2019generative,song2020score} have emerged as a powerful class of deep generative models, achieving state-of-the-art (SOTA) performance in a variety of tasks. These models have gained significant attention due to a variety of compelling reasons. They exhibit remarkable performance in tasks involving (un)conditional image \cite{ho2020denoising,saharia2022image}, audio \cite{kong2020diffwave}, video \cite{ho2022video,he2022latent}, and text \cite{austin2021structured,hoogeboom2021argmax} synthesis and their relatively straightforward implementation compared to other generative models such as Normalizing Flows\cite{rezende2015variational}, GANs\cite{goodfellow2020generative} or VAEs\cite{kingma2013auto}. Furthermore, their connection to Stochastic Differential Equations (SDE) makes the examination of their theoretical properties particularly intriguing \cite{huang2021variational,song2020score,tzen2019neural}. 

Diffusion models were first introduced by Sohl-Dickstein et al., 2015 \cite{sohl2015deep}, draws inspiration from non-equilibrium thermodynamics. The model iteratively destroys the structure in data through a forward diffusion process and then employs a reverse diffusion process to learn how to restore the structure in data. In subsequent work Denoising Diffusion Probabilistic Models (DDPMs) (Ho et al., 2020)\cite{ho2020denoising}, recent advances in deep learning were employed to train a highly effective and adaptable diffusion-based deep generative model, which achieved SOTA results in image synthesis tasks. DDPMs have demonstrated their ability to generate high-fidelity images, often outperforming GANs \cite{dhariwal2021diffusion}, and they achieve this without the need for adversarial training\cite{goodfellow2020generative}. These models, perturb data  by various levels of Gaussian noise in a progressive manner until it  loses its distinguishable features. Samples are then
produced by a Markov chain which, starting from white noise, progressively denoises it into an image. This generative Markov Chain method is either created by reversing a forward diffusion process that gradually converts an image into noise \cite{sohl2015deep} or it is based on Langevin dynamics \cite{song2019generative}.

A notable drawback of diffusion models is their slow sampling time, which often requires thousands of model evaluations, making them unsuitable for real-time applications. Although a recent technique has made progress in reducing the number of necessary sampling steps for diffusion models, their sampling time remains much slower compared to GANs. This slower speed may limit the use of diffusion models in specific applications that prioritize speed and when there are limited computational and memory resources available. To address the issue, in this work, we proposed a knowledge distillation method where the student model is trained to generate samples using a single step.

The motivation of this work is to address the computational intensity of denoising diffusion models, enabling their use on resource-constrained devices. The research aims to integrate denoising phases into the model's architecture, reducing sequential steps and improving efficiency. Additionally, the study seeks to enhance diffusion models by combining them with GANs through knowledge distillation to achieve better framerates during testing.

\section{Background on diffusion models}
\label{sec:bg}
Given a data point sampled from a real data distribution $\mathbf{x}_0 \sim q(\mathbf{x})$, our aim is to learn a model distribution $p_{\theta}(x_{0})$. The parameterized distribution $p_{\theta}(x_{0})$ should closely resemble $q(x_0)$ while also being convenient for generating samples. Denoising diffusion probabilistic models (DDPMs, Ho et al. (2020)\cite{ho2020denoising}) are latent variable models 
with two processes, namely the forward (diffusion) process and the reverse (generative) process. Unlike VAE or flow models,  \cite{rezende2015variational,salimans2015markov}, DDPMs forward process is a non-trainable function and latent variables are relatively high and can be denoted as:
\begin{equation}
\begin{aligned}
\quad
q(\mathbf{x}_{1:T} \vert \mathbf{x}_0) = \prod^T_{t=1} q(\mathbf{x}_t \vert \mathbf{x}_{t-1}) \quad where  \\
\quad
q(\mathbf{x}_t \vert \mathbf{x}_{t-1}) = \mathcal{N}(\mathbf{x}_t; \sqrt{1 - \beta_t} \mathbf{x}_{t-1}, \beta_t\mathbf{I}) 
\end{aligned}
\end{equation}
Where, $x_1, x_2....x_{_T}$ are a sequence of noisy samples. The step size $t$ is controlled by a variance schedule $\beta$ where at each step of the Markov chain, Gaussian noise is added with variance $\{\beta_{t} \in(0,1)\}_{t=1}^T$ producing a new latent variable $x_t$ with transition kernel $q(\mathbf{x}_t \vert \mathbf{x}_{t-1}) $.
As the iteration step $t$ increases, the distinct characteristics of the data sample $\mathbf{x}_0$ gradually fade away. Ultimately, as the value of $T \to \infty$, $x_T$ becomes comparable to an isotropic Gaussian distribution \cite{sohl2015deep}.

\begin{figure}[htb]
\begin{minipage}[b]{1.0\linewidth}
  \centering
  \centerline{\includegraphics[width=9cm]{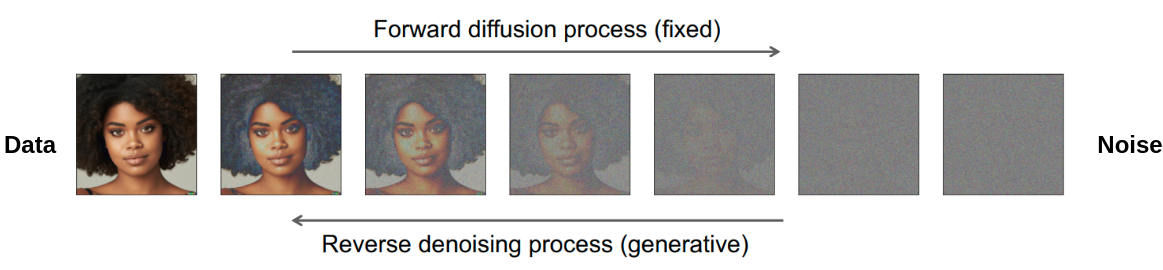}}
\end{minipage}
     \caption{Forward diffusion process.}
\label{fig:fw}
\end{figure}

Figure \ref{fig:fw} shows an illustration of the use of Gaussian diffusion.

The reverse process is estimated by parametrized model as follows:
\begin{equation}
\begin{aligned}
p_\theta(\mathbf{x}_{0:T}) = p(\mathbf{x}_T) \prod^T_{t=1} p_\theta(\mathbf{x}_{t-1} \vert \mathbf{x}_t) \quad where \\ \quad
p_\theta(\mathbf{x}_{t-1} \vert \mathbf{x}_t) = \mathcal{N}(\mathbf{x}_{t-1}; \boldsymbol{\mu}_\theta(\mathbf{x}_t, t), \boldsymbol{\Sigma}_\theta(\mathbf{x}_t, t))
\end{aligned}
\end{equation}

Where, $x_1, x_2....x_{_T}$ are the latent variable with the same dimensionality as $x_0$. 
The parameters $\theta$ are learned to fit the data distribution $q(x_0)$ by maximizing a variational lower bound:
\begin{equation}
\begin{split}
&\max_{\theta} \mathbb{E}_{q(x_0)} \left[ \log p_{\theta}(x_0) \right] \\
&\quad\leq \max_{\theta} \mathbb{E}_{q(x_0,x_1,...,x_T)} \left[ \log p_{\theta}(x_{0:T}) - \log q(x_{1:T} | x_0) \right]
\end{split}
\label{elbo}
\end{equation}
where $q(x_{1:T} | x_0)$ is some inference distribution over the latent variables. 

\begin{figure}[htb]
\begin{minipage}[b]{1.0\linewidth}
  \centering
  \centerline{\includegraphics[width=9cm]{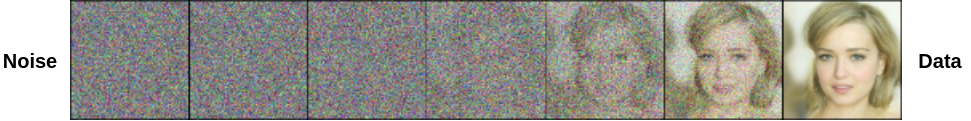}}
\end{minipage}
    \caption{Reverse denoising process.}
\label{fig:rv}
\end{figure}

A special property of the forward process is that we can sample $x_t$
 at any arbitrary time step $t$ in a closed form using reparameterization trick \cite{kingma2013auto}. So $x_t$ can be expressed as a linear combination of $x_0$ and a noise variable $\epsilon$:
 
 \begin{equation}
 \begin{aligned}
\mathbf{x}_t &= \sqrt{\alpha_t}\mathbf{x}_{0} + \sqrt{1 - \alpha_t}\boldsymbol{\epsilon} & \text{ where } \boldsymbol{\epsilon} \sim \mathcal{N}(\mathbf{0}, \mathbf{I}) \\ 
 \end{aligned}
 \end{equation}
If all the conditionals are modeled as Gaussians with trainable
mean functions and fixed variances, the objective in Eq \ref{elbo}can be simplified to \footnote {Please refer to \cite{weng2021diffusion} for details}:

\begin{equation}
    \begin{aligned}
L_t^\text{simple}
&= \mathbb{E}_{t \sim [1, T], \mathbf{x}_0, \boldsymbol{\epsilon}_t} \Big[\|\boldsymbol{\epsilon}_t - \boldsymbol{\epsilon}_\theta(\mathbf{x}_t, t)\|^2 \Big] \\
&= \mathbb{E}_{t \sim [1, T], \mathbf{x}_0, \boldsymbol{\epsilon}_t} \Big[\|\boldsymbol{\epsilon}_t - \boldsymbol{\epsilon}_\theta(\sqrt{\bar{\alpha}_t}\mathbf{x}_0 + \sqrt{1 - \bar{\alpha}_t}\boldsymbol{\epsilon}_t, t)\|^2 \Big]
\end{aligned}
\end{equation}

In DDPMs, the length of the forward process $T$ is a crucial hyperparameter. The selection of large $T$ values, such as $T = 1000 $ \cite{ho2020denoising}, is driven by the fact that, from a variational perspective, a large $T$ enables the reverse process to be nearly Gaussian process\cite{sohl2015deep}, making the generative process modeled with Gaussian conditional distributions a good approximation. However, sampling from DDPMs is slower than sampling from other deep generative models since all $T$ iterations must be carried out sequentially rather than in parallel to acquire a sample $x_0$, making them unsuitable for tasks where computational resources are constrained, and quick response times are essential. 

The sampling steps of DDPMs can be written as 

$x_0=f(f(f...f(x_{_T})...))$
Where $x_{_T}$ is a normally distributed noise with a mean of zero and unit variance, $x_0$ is the final image from the model and $f$ is the non-linear transformation applied by the model. As it can be seen from the equation, first the model is applied to $x_{_T}$, and then its applied to the previous outputs of the model until we reach $x_0$. It should be noted that all ${x_0, x_1, ..., x_{_T}}$ lie in the same high dimensional space. This increases the number of parameters required by the   diffusion models  making them  slow during  sampling.

This study introduces two modifications to the diffusion model. Instead of utilizing the identical model at each step during the backward diffusion process, we employed different models at each step. The equations employed is as follows:
$x_0=f_0(f_1(f_2...f_{_T}(x_{_T})...))$.
Here, we refer to these distinct models utilized at each step as submodules, denoted by the functions $f_i$. These submodules are composed of few non-linear layers. In addition, we started with low dimensional noise ($x_{_T}$) and gradually increased the resolution of the intermediate outputs until we reach $x_0$. Consequently, this enabled us to reduce the number of parameters required for the model.

In this study, we developed a compact student model aimed at approximating the DDPM's output, achieved by minimizing the steps needed to generate an image. Unlike the teacher model which employs thousands of steps, the student model produces images in just one step. Our objective was to guide the student model's submodules using intermediate outputs from the teacher model. This involved using downsampled teacher outputs as points of guidance. For this purpose, we employed GANs, utilizing teacher outputs as actual samples and student outputs as fake samples, effectively supervising the submodules' training process. The ultimate goal is to find a method that enables denoising to be achieved in a single step, thus reducing the total number of steps required while maintaining a high quality and fidelity image quality. By enhancing the efficiency of these models, the intention was to address the resource-intensive nature of the models and make them more feasible for real-world applications in scenarios with limited resources.

\section{Adversarial Knowledge Distillation}
\label{sec:approach}
In this work, we utilized a pre-trained diffusion teacher model which often takes many steps to sample images, and train a student model that can generate images with a single pass. We integrated the diffusion process into the student model architecture and downsampled versions of $x_t$ from the teacher model are used to supervise the output from submodules of the student model. 

\begin{figure}[htb]
\begin{minipage}[b]{1.0\linewidth}
  \centering
  \centerline{\includegraphics[width=8.5cm]{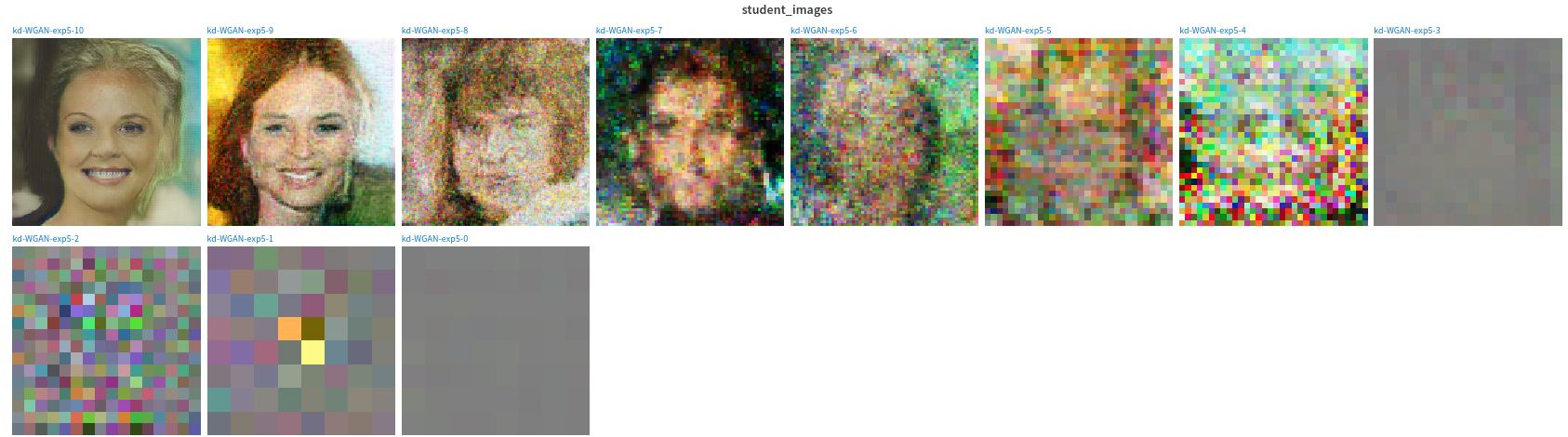}}
\end{minipage}
    \caption{Outputs from different submodules of the student model. }
\label{fig:stud}
\end{figure}

Our model draws inspiration from the success of deep generative models in converting noise inputs into deep features \cite{karras2019style}. Accordingly, our model employs a similar technique by transforming normally distributed noise into a facial image and performing denoising operations in successive layers.
Figure \ref{fig:stud} illustrates the outputs from different submodules of the student model. Each submodule applies a denoising step and even-numbered submodules upsample their inputs.

\begin{figure}[htb]
\begin{minipage}[b]{1.0\linewidth}
  \centering
  \centerline{\includegraphics[width=8.5cm]{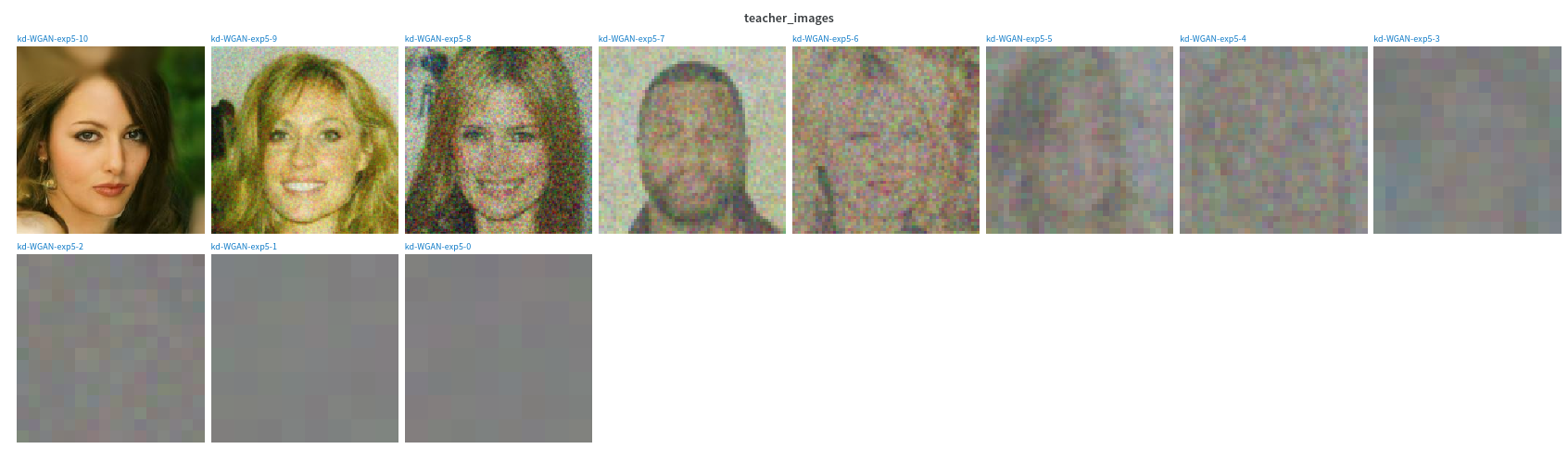}}
\end{minipage}
    \caption{Outputs from different time steps of the teacher model (DDPM).}
\label{fig:tech}
\end{figure}

The teacher transforms a normally distributed noise $X_T$ into a face image($x_0$) by sequentially applying the same model again and again, for $T$ steps.
Figure \ref{fig:tech} illustrates the outputs of different time steps of the teacher model.

Let $X$ be a set of outputs from the teacher model $(X = x_{_{T-1}}, x_{_{T-2}}, ..., x_1, x_0)$. The downsampled version of $X$ is given by $(X' = x'_{_{T-1}}, x'_{_{T-2}}, ..., x'_1, x'_0)$ where  $x'_t = \Psi(x_{_t}, t)$. 
\\We used the downsampled versions of the teacher model outputs to supervise the student model. Here  $\Psi(x_{_t}, t)$ is a downsampling function and depends on the time $t$. 

The student model is composed of submodules and the submodules are composed of a few non-linear transformation layers.  
The student model can be denoted as 
\begin{equation}
\begin{aligned}
y_{_0} = f(y_{_T}) = f_0(f_1...(f_{T-1}(y_{_T}))..) \quad 
and \quad y_{_t} = f_t(y_{_{t+1}}) 
\end{aligned}
\end{equation}
where each $f_t$ is a submodule of the student model and  $y_{_t}$ is output from the submodule.\\
Since $T$ is often large, creating a submodule for each timestep makes the student model very huge. Rather than creating the submodules for all time steps,  in this work, only  $11$  timesteps are chosen from the space of all timesteps and the student model has only  $11$ internal submodules.
\begin{equation}
\begin{aligned}
 y_{_n} = f_n(y_{_{n+1}})
\end{aligned}
\end{equation}
where  $n \in {0, 1, 2, 3,..., 10} $ and $y_{_n}$ corresponds to teacher's output at timestep $t = 190n - 10n^2 + 80 $ for all $n$ except $n=11$. For $n = 11$, the last output from the teacher model is selected.

\begin{figure}[htb]
\begin{minipage}[b]{1.0\linewidth}
  \centering
  \centerline{\includegraphics[width=8.5cm]{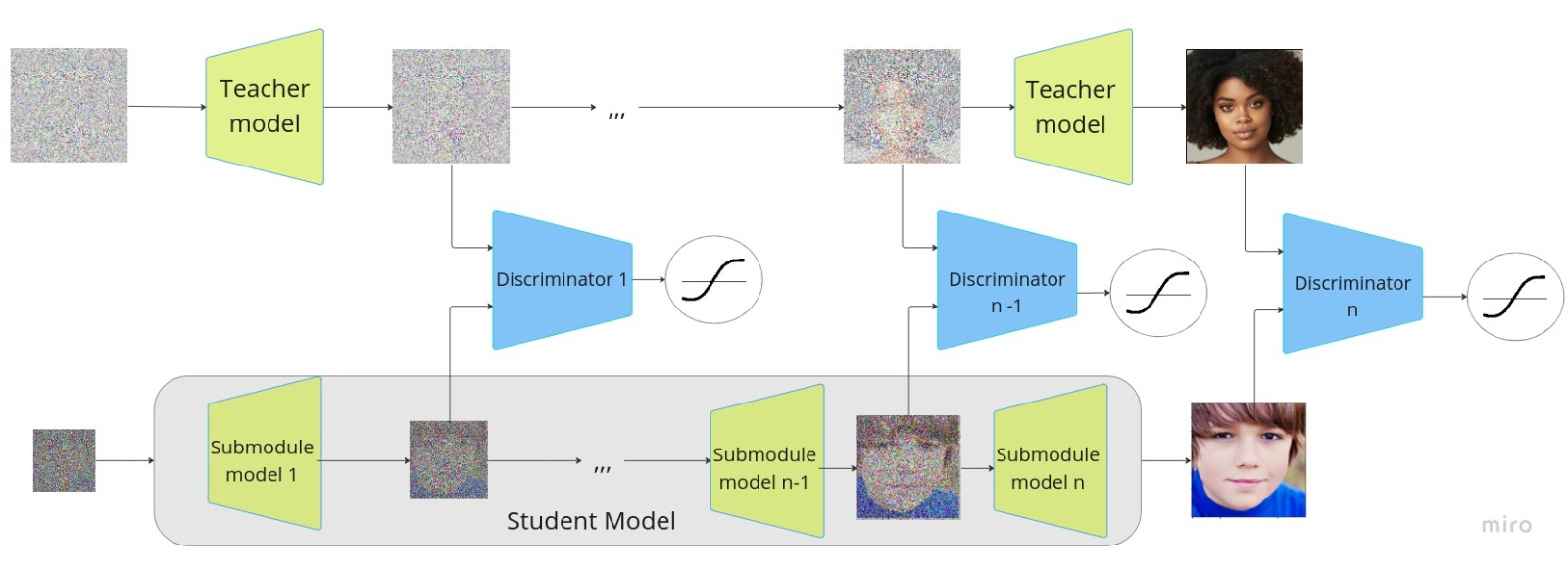}}
\end{minipage}
    \caption{Knowledge distillation using adversarial training.}
\label{fig:arc}
\end{figure}

Figure \ref{fig:arc}illustrates Knowledge distillation using adversarial training. The teacher model starts from a noise image and tries to denoise it at each step. While the student model internally applies the denoising step as well as up-samples the input. The discriminator is applied to the intermediate outputs.

The objective function for the student model now can be written as:-
\begin{equation}
\begin{aligned}
 \mathcal{L}_n =\mathcal{J}	(y_{_{n}}, x'_{_t}) 
\end{aligned}
\end{equation}
 where  $x'_{_t}$  is a downsampled image from the teacher model at step $t= 190n - 10n^2 + 80$. 
Any differentiable distance function can be utilized as the loss function $\mathcal{J}(,)$, such as Mean Squared Error Loss, KL Divergence Loss, or Wesserstein (Earth Mover) Loss. However, in this study, we opted to use the adversarial training loss known as Wasserstein Loss, and is formulated as :
\begin{equation}
\begin{aligned}
 L(D^{n}_\theta, G^{n}_\theta) = \mathbb{E}_xP_r[D^{n}_\theta(x_{_{t'}})] - \mathbb{E}_zP_z[D^{n}_\theta(G^{n}_\theta(y_{_{n+1}}))]
\end{aligned}
\end{equation}
Where $G^{n}_\theta$ is $n^{th}$  submodule of the student model, and $D^{n}_\theta$ is discriminator at step $n$.

\section{EXPERIMENTS}
\label{sec:experiment}

In this section, we empirically validate the adversarial distillation method  proposed in Section\ref{sec:approach},
\subsection{Experimental Data and Dataset Description.}

\textbf{CelebA Dataset}:- 
Our teacher models are pretrained on the CelebA dataset\cite{liu2015faceattributes}, a large-scale face attributes dataset with over 200,000 celebrity face images. We have randomly selected 1000 images from CelebA to evaluate our teacher and student model.

\textbf{DDPM  generated dataset}:- 
The main dataset used to train our student models is generated from a pretrained diffusion teacher model. We have generated (256x256) images using the DDPM model. For each image, we sampled noise inputs from a standard normal distribution and applied the model to the noise to get face images.  In addition to the final outputs, we have saved the inputs and intermediate outputs of the teacher model which are used to supervise the student model. We have selected intermediate outputs which are generated at timesteps that are multiple of 100. The dataset comprises 20,000 data points generated using this approach.

\textbf{DDIM generated dataset} :- 
Similar to the DDPM generated dataset, we used the DDIM pretrained model as a teacher model and generated $\sim 11,000$ cases where each case has noise inputs to the teacher model, intermediate outputs from the teacher model at some predetermined steps, and outputs from the teacher model. We use this dataset to train another student model. This teacher model also uses $1000$ timesteps to generate samples, but it performs inference on every 20 timesteps for faster sampling resulting in 50 total steps. From these timesteps, we selected only $11$ steps to train the student model. To determine the choice of the timesteps we used a decreasing quadratic scale, where the student submodules learned to predict images at timesteps $t$ determined by a quadratic function. In this case, for $n$ in ${0, 1, 2,\ldots, 10}$ we chose $190n - 10n^2 + 80$ timesteps.

The student model is a generative model motivated by the DCGAN\cite{radford2015unsupervised} model. In addition, we also created a discriminator model similar to \cite{radford2015unsupervised}.
The student model consists of $12$ submodules, with $11$ of them utilized for predicting intermediate outputs, while the final submodule is designed to generate image samples. Each submodule is comprised of two convolutional layers followed by batch normalization and Leaky ReLU activation. Furthermore, each submodule is linked to a terminal module that maps the features to images for training purposes, but these terminal modules do not form part of the final model. Consequently, the student model has significantly decreased in size, measuring around $9$MB, in contrast to the original DDPM model, which had a size of approximately $455$MB.

During the optimization process, we employed varying batch sizes for different submodules. Specifically, for submodules $1$ through $8$, a batch size of $64$  was utilized, whereas for the $9^{th}$ submodule, a batch size of $32$ was used. For the remaining submodules, a batch size of $16$ was employed. Our decision on which batch size to use was based on the number of parameters being optimized.

\subsection{Results}
\label{sec:res}
The teacher model and student model were evaluated based on their FID scores, which provide a quantitative measure of the performance of generative models. Lower FID scores indicate better model performance. To obtain FID scores for both models, we randomly selected 1000 images from the CelebA dataset, student model samples, and teacher model samples. 

Table \ref{tab:table1} presents the FID scores for both models. These results indicate a decrease in performance when training smaller student models to approximate the outputs of the teacher model. However, compared to the sizes of the models, the decrease in FID score of the student model is minor. 

\begin{table}[htbp]
\begin{center}
 \begin{tabular}{ |c|c| } 
 \hline
 Model Name & FID  Score \\
\hline
 Teacher model & 60.9978 \\ 
 \hline
 Student model & 142.5443\\ 
  \hline
\end{tabular}
\end{center}
  \caption{Results of the student model as compared to the teacher DDPM model\label{tab:table1}}
\end{table}

The obtained results reveal a noticeable difference between the FID scores of the student and teacher models. This discrepancy highlights a noticeable decline in performance when trying to train smaller student models to approximate the outcomes produced by the larger teacher model. However, when accounting for the contrasting sizes of these models, the reduction in the FID score of the student model appears relatively modest. This observation stems from the fact that our model boasts approximately 50 times fewer parameters, requiring substantially less memory. Despite these resource advantages, the student model adeptly succeeds in approximating the intricate outputs of the teacher model.

\begin{figure}[htb]
\begin{minipage}[b]{1.0\linewidth}
  \centering
  \centerline{\includegraphics[width=8.5cm]{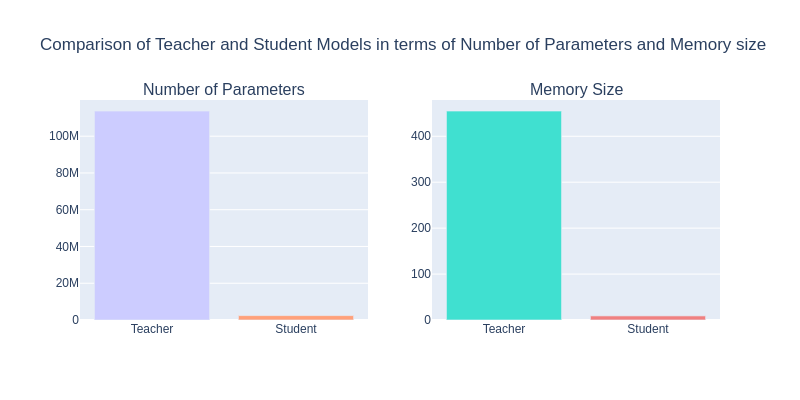}}
\end{minipage}
\label{fig:size}
\end{figure}

\begin{table}[htbp]
\begin{center}
\begin{tabular}{ |c|c|c| } 
\hline
 Model Name  & Number of Parameters   &  Memory Size\\
\hline
 Teacher model & 113.7 M & 455 MB\\
\hline
 Student model & 2.4 M & 9 MB \\
  \hline
\end{tabular}
\end{center}
\caption{Comparison of the model sizes of the student model and the teacher model\label{tab:table2}}
\end{table}

\begin{figure}[htb]
\begin{minipage}[b]{1.0\linewidth}
  \centerline{\includegraphics[width=8.5cm]{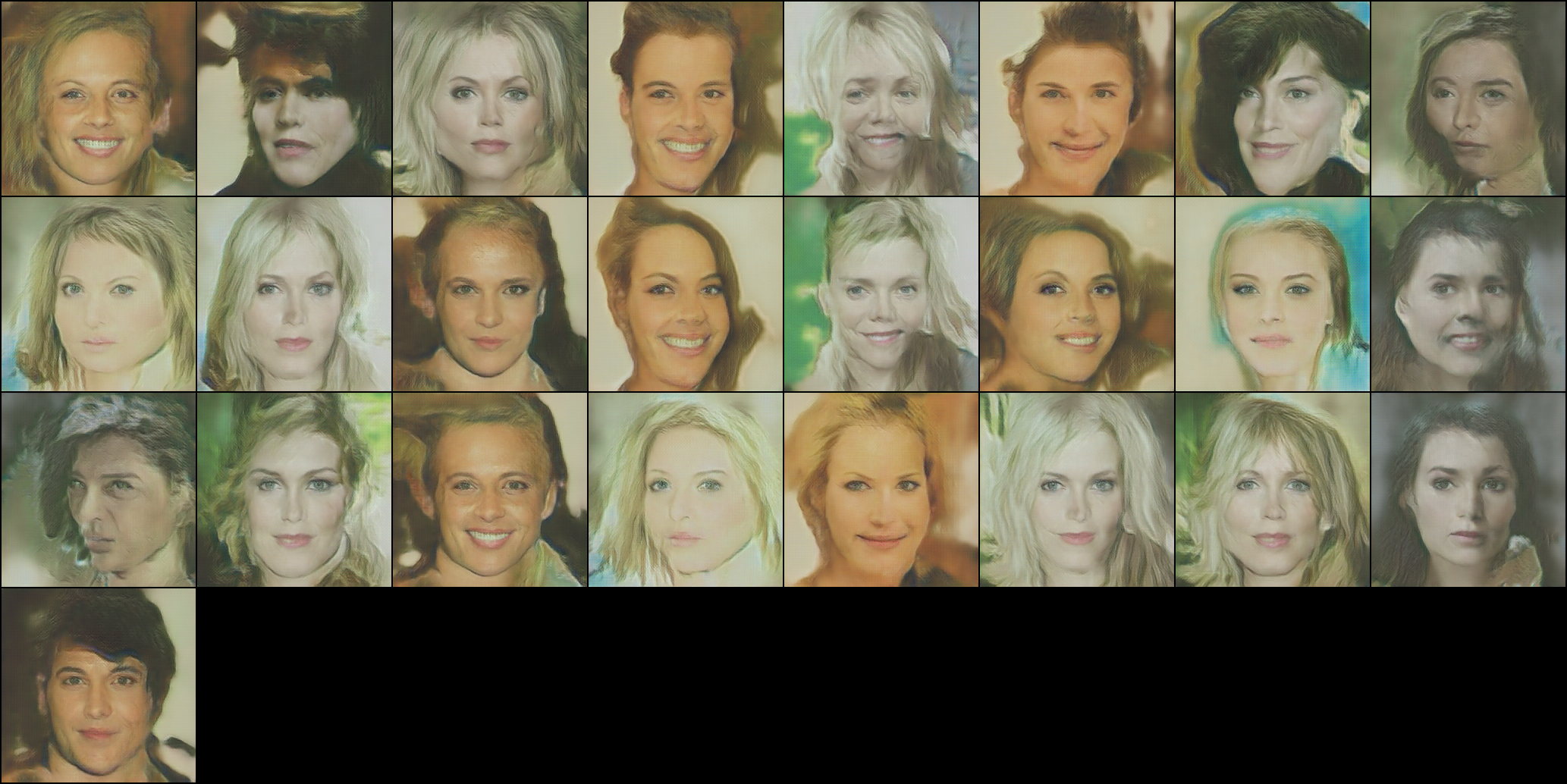}}
\end{minipage}
  \caption{Sample images generated by the student model.}
    \label{fig:studs}
\end{figure}

\begin{figure}[htb]
\begin{minipage}[b]{1.0\linewidth}
  \centerline{\includegraphics[width=8.5cm]{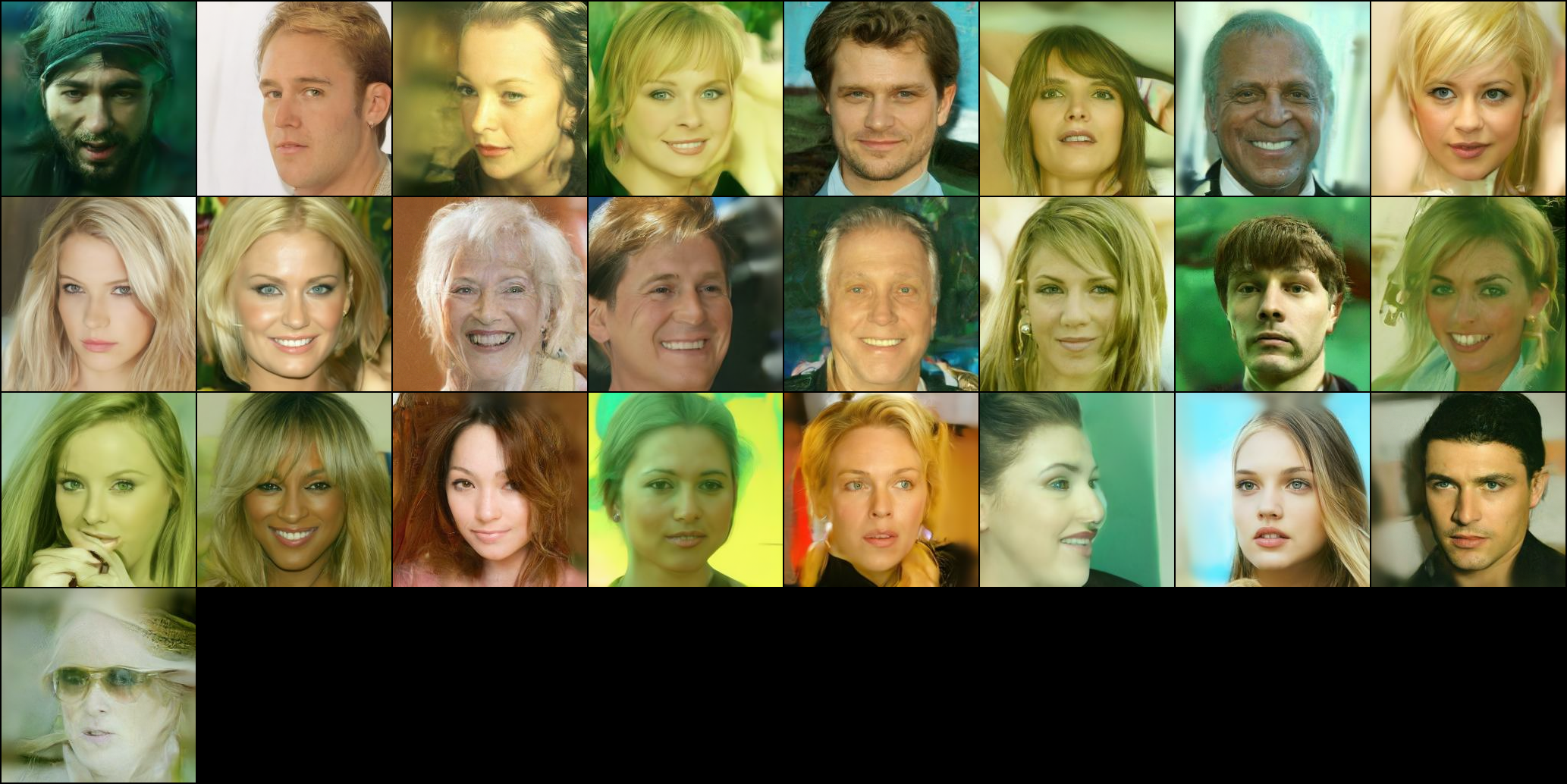}}
\end{minipage}
   \caption{Sample images generated by the teacher model.}
    \label{fig:techs}
\end{figure}

Figure \ref{fig:studs} and \ref{fig:techs} illustrate the sample images generated by the student and teacher models respectively.

\section{Related Work}
\label{sec:related-work}
Our research builds upon a large family of existing methods for learning generative models. Several generative models that use diffusion as a basis have been proposed in scholarly literature, The three predominant formulations of diffusion models are Denoising diffusion probabilistic models (DDPMs) \cite{sohl2015deep,ho2020denoising}, score-based generative models (SGMs)\cite{song2019generative,song2020improved}, and stochastic differential equations (Score SDEs)\cite{karras2022elucidating,song2021maximum,song2020score}. These models all possess the progressive perturbation of data with intensifying levels of random noise (known as the "diffusion" process) to the data, followed by the removal of noise to generate new data samples. 

Multiple approaches have been developed to improve the performance of diffusion models. These methods aim to either improve the models' performance by increasing their empirical efficiency and accuracy in describing real-world data through more efficient algorithms and techniques \cite{nichol2021improved,song2020denoising,song2020improved} or to make the models more versatile and applicable to a wider range of scenarios by broadening their theoretical scope and incorporating new factors \cite{lu2022maximum,song2020score,lu2022dpm,song2021maximum,zhang2022fast}.
\\
Our suggested approach closely aligns with the research conducted by (Salimans et al., 2022) \cite{salimans2022progressive}. This method involves reducing the number of steps required for sampling without sacrificing sample quality, by introducing new parameterizations to increase stability and a distillation technique that transforms a diffusion model with N sampling steps into a new model with N/2 steps with minimal degradation in sample quality. To improve the quality of samples that may be affected by the use of fewer steps in the sampling process, the authors propose using new parameterizations for diffusion models and new weighting schemes in the objective function. However, this method requires progressive training from fine to coarse resolution and may reach the limit of a function evaluation.

\cite {luhman2021knowledge} is another  approach closely resembling our proposed technique. The author engages in the distillation of DDIM teacher models into one-step student models. However, there's a potential drawback to their approach as it necessitates the creation of an extensive dataset through the complete execution of the original model at its maximum sampling steps. Consequently, the expense associated with their distillation method increases linearly with the number of these steps, which might be restrictive. In contrast, our method eliminates the requirement of executing the original model with its full sampling steps. Only 11 timesteps are selected out of the teacher's 1000 steps.

Song et al. (2020) \cite{song2020denoising} introduced denoising
diffusion implicit models (DDIMs), a more efficient class of iterative implicit probabilistic models that
are based on a generalized version of DDPMs. DDIMs are defined by a class of non-Markovian diffusion
processes that can represent deterministic generative processes and use the same training procedure as
DDPMs.  While DDIMs present a novel approach to enhancing the efficiency of diffusion probabilistic models, it is important to note that our proposed method takes a different avenue to address similar challenges. In our approach, we integrate denoising phases directly into the model's architecture and leverage generative adversarial networks (GANs) through knowledge distillation. By combining these techniques, we achieve reduced computational requirements and improved sampling speed without altering the fundamental probabilistic model.

Meng et al, 2022) \cite{meng2022distillation}, address the computational expense of classifier-free guided diffusion models \cite{ho2022classifier} at inference time by proposing a method for distilling them into models that are faster to sample from. To do this, they first learn a single model to match the output of a combined conditional and unconditional model, and then gradually distill that model into a diffusion model that requires significantly fewer sampling steps using a pre-trained classifier-free guided model. The authors demonstrate that their approach can generate high-quality images using fewer sampling steps than the original model, making it faster and more efficient for inference. This approach is effective for standard diffusion models trained on the pixel space and diffusion models trained on the latent space and can be applied to tasks such as text-guided image editing and inpainting.
This contribution resonates with the goals of our research, which focuses on improving the efficiency of diffusion models through the integration of denoising phases. While our approach takes a different architectural route, this work holds the potential for influencing our own method. Furthermore, the significant acceleration in sampling speed achieved by the distilled diffusion model aligns with our aim to mitigate the computational burden associated with iterative denoising steps.

\section{Conclusion \& Future Work}
\label{sec:conclusion}

In this study, we have proposed a new approach that incorporates diffusion steps into the model architecture. Our method involves training compact student diffusion model using knowledge distillation techniques, enabling the generation of high-quality sample images in a single internal denoising step, thereby eliminating the requirement for consecutive denoising procedures. Adversarial training optimizes the student model by treating teacher outputs as real images and student outputs as fake during training, building upon a pretrained DDPM model.

\begin{figure}[htb]
\begin{minipage}[b]{1.0\linewidth}
  \centerline{\includegraphics[width=8.5cm]{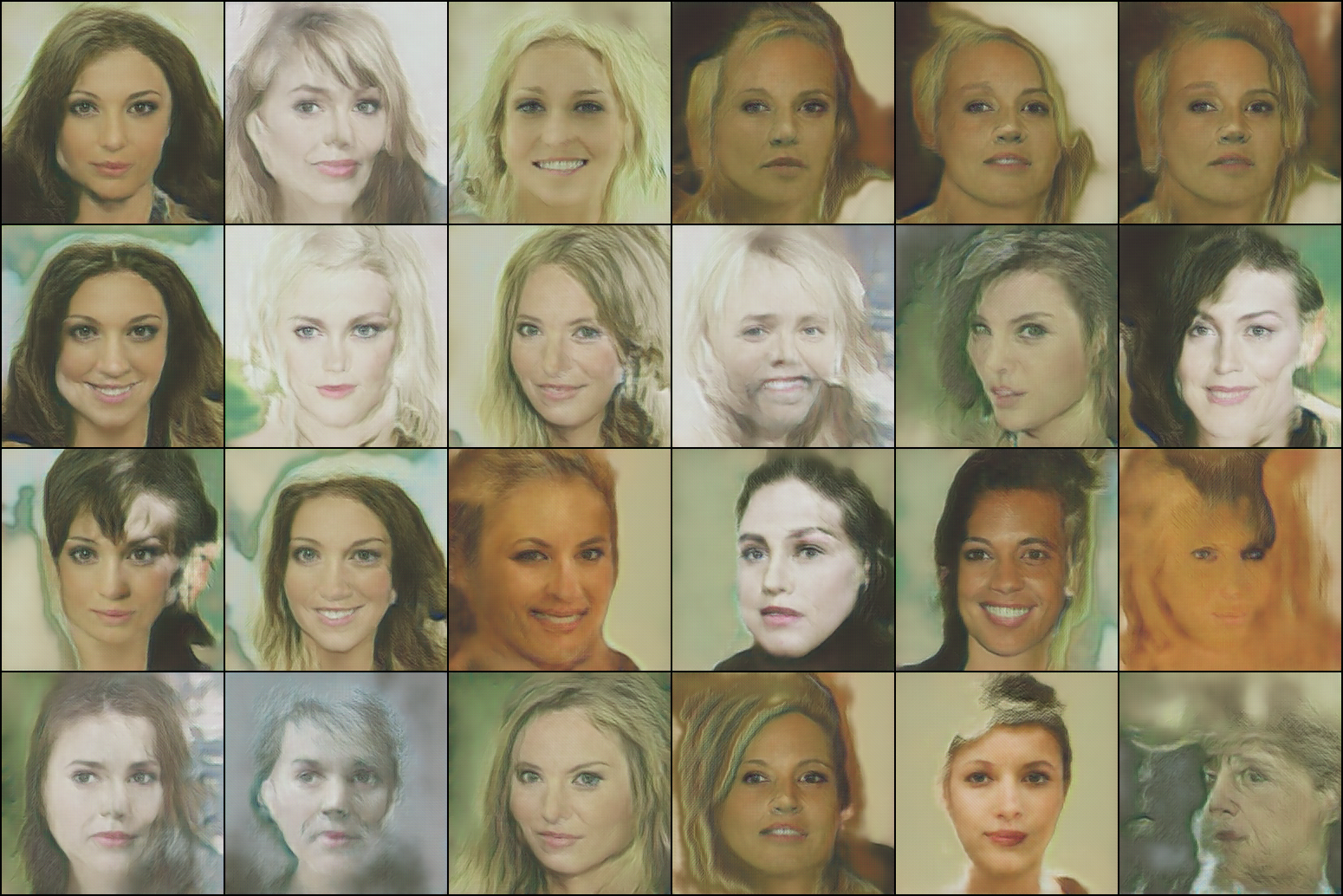}}
\end{minipage}
  \caption{Student model sample diversity issue.}
    \label{fig:divers}
\end{figure}

\begin{figure}[htb]
\begin{minipage}[b]{1.0\linewidth}
  \centerline{\includegraphics[width=8.5cm]{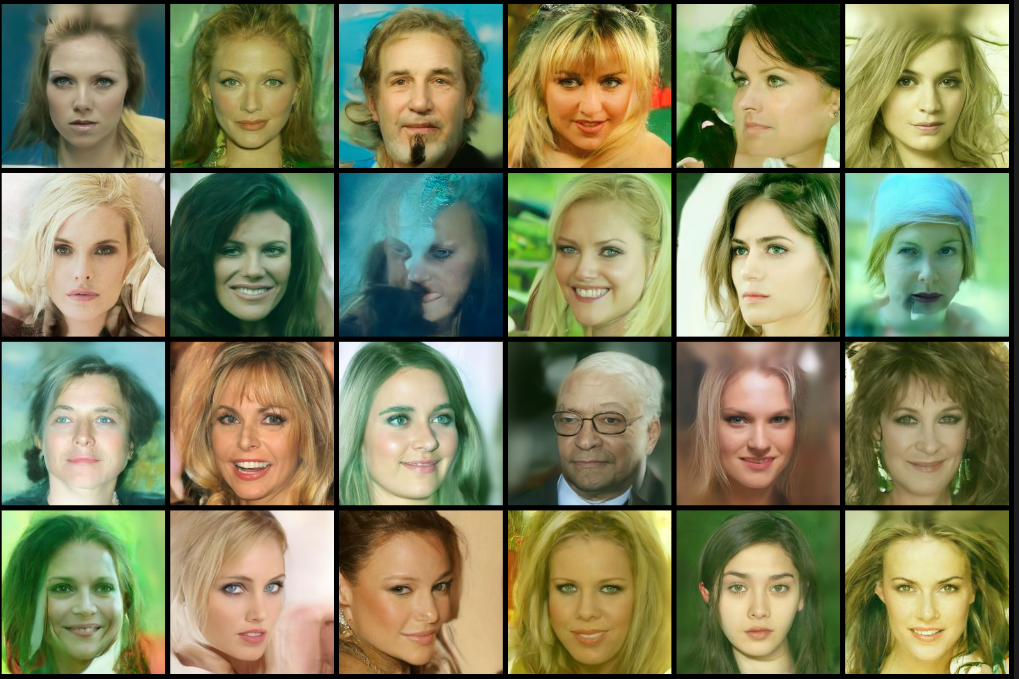}}
\end{minipage}
  \caption{The problem of image deformation in the teacher's sample can be observed in the images of both the first and second rows.}
    \label{fig:teach_distor}
\end{figure}

\begin{figure}[htb]
\begin{minipage}[b]{1.0\linewidth}
  \centerline{\includegraphics[width=8.5cm]{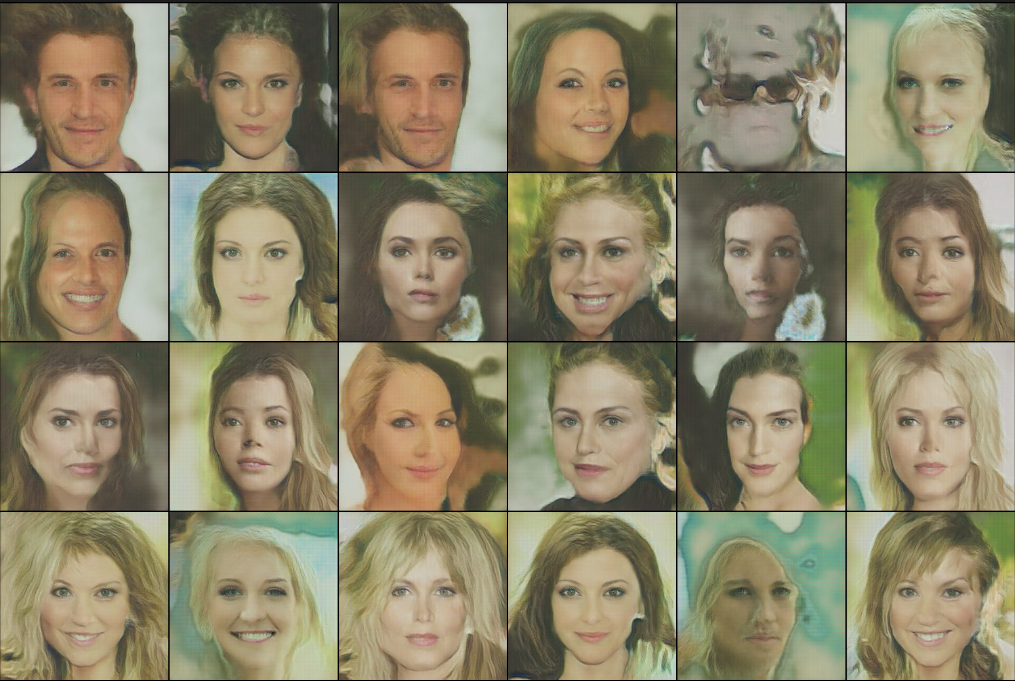}}
\end{minipage}
  \caption{Student’s sample images deformation issue.}
    \label{fig:distor}
\end{figure}

Notably, the student model demonstrates its capacity to generate high-resolution images  with fewer parameters  and carry out internal denoising, in contrast to the stepwise denoising procedure inherent in the teacher model.  This approach eliminates the need for consecutive denoising steps, providing a practical solution for resource-constrained real-world scenarios. However, it's important to recognize that the GAN-trained student model may display reduced image diversity \ref{fig:divers} compared to diffusion models (as shown in Figure \ref{fig:divers}), 
and similar to the characteristics of the teacher model (as depicted in Figure \ref{fig:teach_distor}), the student model might display minor distortions in the generated images (as illustrated in Figure \ref{fig:distor}).

To mitigate these limitations and enhance both the diversity and fidelity of the generated images, we propose several directions for future research. One avenue involves exploring alternative DDPM models that could potentially address the observed reduction in image diversity. Additionally, augmenting the dataset used in our study is a promising approach. Currently, our investigation relied on a dataset comprising 20,000 facial images generated by the teacher model (DDPM). Expanding this dataset to incorporate a more extensive array of images generated by the teacher model could potentially yield improved results. This expanded dataset would afford the student model the opportunity to learn from a broader spectrum of data, likely resulting in enhanced performance.

Moreover, we recommend considering alternative loss functions, such as mean squared error or KL-Divergence, for training the student model when equipped with a larger dataset. These alternate loss functions could potentially alleviate the issue of image deformation observed in our study. While these avenues remain unexplored within the scope of this research, we recognize their potential value and have designated them as subjects for future investigation.

\section{ETHICS STATEMENT}
\label{sec:ethics}
Typically, generative models can be vulnerable to unethical applications, including the generation of deceptive content and the potential for bias when used with insufficiently curated datasets. The core objective of this paper centers on the optimization of generative models for efficient testing, aiming to alleviate potential computational burdens while remaining mindful of ethical considerations.

\section{Acknowledgment}
This paper, authored by Mekonnen Kidist Amde, was initially completed as part of the requirements for obtaining an MSc in Data Science from the University of Trento, Italy, within the Department of Mathematics, during the academic year 2021/2022. The thesis, now titled "ADV-KD: ADVERSARIAL KNOWLEDGE DISTILLATION FOR FASTER DIFFUSION SAMPLING," represents an evolution of the original research. It was formally presented on March 8, 2023, as documented in the final dissertation available on the school website's Biblioteca (\href{https://www5.unitn.it/Biblioteca/en/Web/TesiDocente/192681}{Link}). This work was also presented at the \href{https://fcai.fi/eds2023/review}{ELLIS Doctoral Symposium 2023 in Helsinki, Finland}. The author extends heartfelt gratitude to the supervisor, Paolo Rota, and co-supervisor, Nicola Dall'Asen, for their invaluable support and guidance.

\bibliographystyle{IEEEbib}
\bibliography{strings,refs}

\end{document}